\documentclass{article}
\PassOptionsToPackage{numbers, compress}{natbib}
\usepackage[preprint]{nips_2018}
\usepackage{hyperref}
\usepackage{url}
\usepackage{amssymb}
\usepackage{booktabs} %需要加载宏包{booktabs}
\usepackage{algorithm}
\usepackage{algpseudocode}
\usepackage{amsfonts}
\usepackage[marginal]{footmisc}
\usepackage{amsmath}
\usepackage{epsfig}
\usepackage{graphicx}
\usepackage{epstopdf}
\usepackage{subfigure}
\usepackage{hyperref}
\usepackage[utf8]{inputenc} % allow utf-8 input
\usepackage[T1]{fontenc}    % use 8-bit T1 fonts
\usepackage{hyperref}       % hyperlinks
\usepackage{url}            % simple URL typesetting
\usepackage{booktabs}       % professional-quality tables
\usepackage{amsfonts}       % blackboard math symbols
\usepackage{nicefrac}       % compact symbols for 1/2, etc.
\usepackage{microtype}      % microtypography

\title{Unsupervised/Semi-supervised Deep Learning for \\
Low-dose CT Enhancement
}

\author{
  Mingrui Geng\thanks{Mingrui Geng and Yun Deng made equal contributions to this work.}\\
  Xian Jiaotong University\\
  \And
  Yun Deng\\
  Xian Jiaotong University \\
  \And
  Qian Zhao \\
  Xian Jiaotong University \\
  \AND
  Qi Xie \\
  Xian Jiaotong University \\
  \And
  Dong Zeng \\
  Southern Medical University \\
  \And
  Jianhua Ma \\
  Southern Medical University \\
  \AND
  Wangmeng Zuo \\
  Harbin Institute of Technology, China \\
  \And
  Deyu Meng \thanks{Deyu Meng is the corresponding author. Email: dymeng@mail.xjtu.edu.cn.}\\
  Xian Jiaotong University \\
}

\begin{document}

\maketitle

\vspace{-4mm}

\begin{abstract}\vspace{-1mm}
Recently, deep learning (DL) methods have been proposed for the low-dose computed tomography (LdCT) enhancement, and obtain good trade-off between computational efficiency and image quality. Most of them need to pre-collect large number of ground-truth/high-dose sinograms with less noise, and train the network in a supervised end-to-end manner. This may bring major limitations on these methods because the number of such low-dose/high-dose training sinogram pairs would affect the network's capability and sometimes the ground-truth sinograms are hard to be obtained in large scale. Since large number of low-dose ones are relatively easy to obtain, it should be critical to make these sources play roles in network training in an unsupervised learning manner. To address this issue, we propose an unsupervised DL method for LdCT enhancement that incorporates unlabeled LdCT sinograms directly into the network training. The proposed method effectively considers the structure characteristics and noise distribution in the measured LdCT sinogram, and then learns the proper gradient of the LdCT sinogram in a pure unsupervised manner. Similar to the labeled ground-truth, the gradient information in the unlabeled LdCT sinogram can be used for sufficient network training. The experiments on the patient data show effectiveness of the proposed method.

\end{abstract}

\vspace{-4mm}
\section{Introduction}\vspace{-2mm}
Computed tomography (CT) has been widely used for clinical diagnosis.
Meanwhile, concerns regarding radiation-related cancer in CT examination are growing, especially in the repeated CT scans \cite{Brenner2016MO}. Therefore, decreasing X-ray dose to reduce risk to patients is highly desired. However, this would lead to severe noise-induced artifacts in the filtered backprojection (FBP) reconstructed image without adequate treatments \cite{Hsieh1998Adaptive}\cite{Lu2001Noise}\cite{Xu2009Electronic}.

Many methods have been proposed to enhance the LdCT image quality. These methods can be mainly divided into two strategies. One is to characterize noise distribution or designate somewhat handcrafted prior based on the conventional maximum a posteriori probability (MAP) model \cite{Ma2012Iterative}\cite{Ouyang2011Effects}\cite{Wang2009Iterative}\cite{Xie2017Robust}. Although these MAP-based methods can yield high-quality LdCT sinograms to some extent, they may have intrinsic limitations: First, the iterative solution process of these methods yields a high computational cost, and can be hundreds of times slower than DL methods in prediction process. Second, these methods process each sinogram separately, and thus are not able to integrate all CT sinogram sources to extract their common latent knowledge underlying desired CT sinograms.

The other is deep learning (DL) approach, which learns the mapping from the LdCT images to high-dose ones in an end-to-end manner \cite{Chen2017Low}\cite{Chen2016Low}\cite{Kang2017A}\cite{Kenji2017Neural}, and has obtained state-of-the-art performance on the task. However, this line of methods needs to pre-collect a large quantity of low-dose/high-dose CT image pairs as the training inputs/outputs of the network, where the labeled LdCT images are usually generated from the high-dose images via simulation methods. However, due to the limitations of privacy, collecting costs and domain biases, it is always impractical to attain sufficient training sample pairs as expected. Moreover, the current CT image enhancement methods based on DL do not take good advantage of abundant information in the unlabeled CT dataset.

It thus has become a critical issue to make unsupervised LdCT images, without guidance of the corresponding high-dose ones, capable of being sufficiently involved in deep network training. Such an unsupervised deep learning issue actually has been attracting increasing attention throughout machine learning \cite{Lehtinen2018Noise2Noise}, pattern recognition \cite{Lotter2016Deep}, computer vision \cite{Yin2018GeoNet}, and many other related domains.

\begin{figure}\vspace{-4mm}
	\begin{center}
		\includegraphics[width=0.78\textwidth]{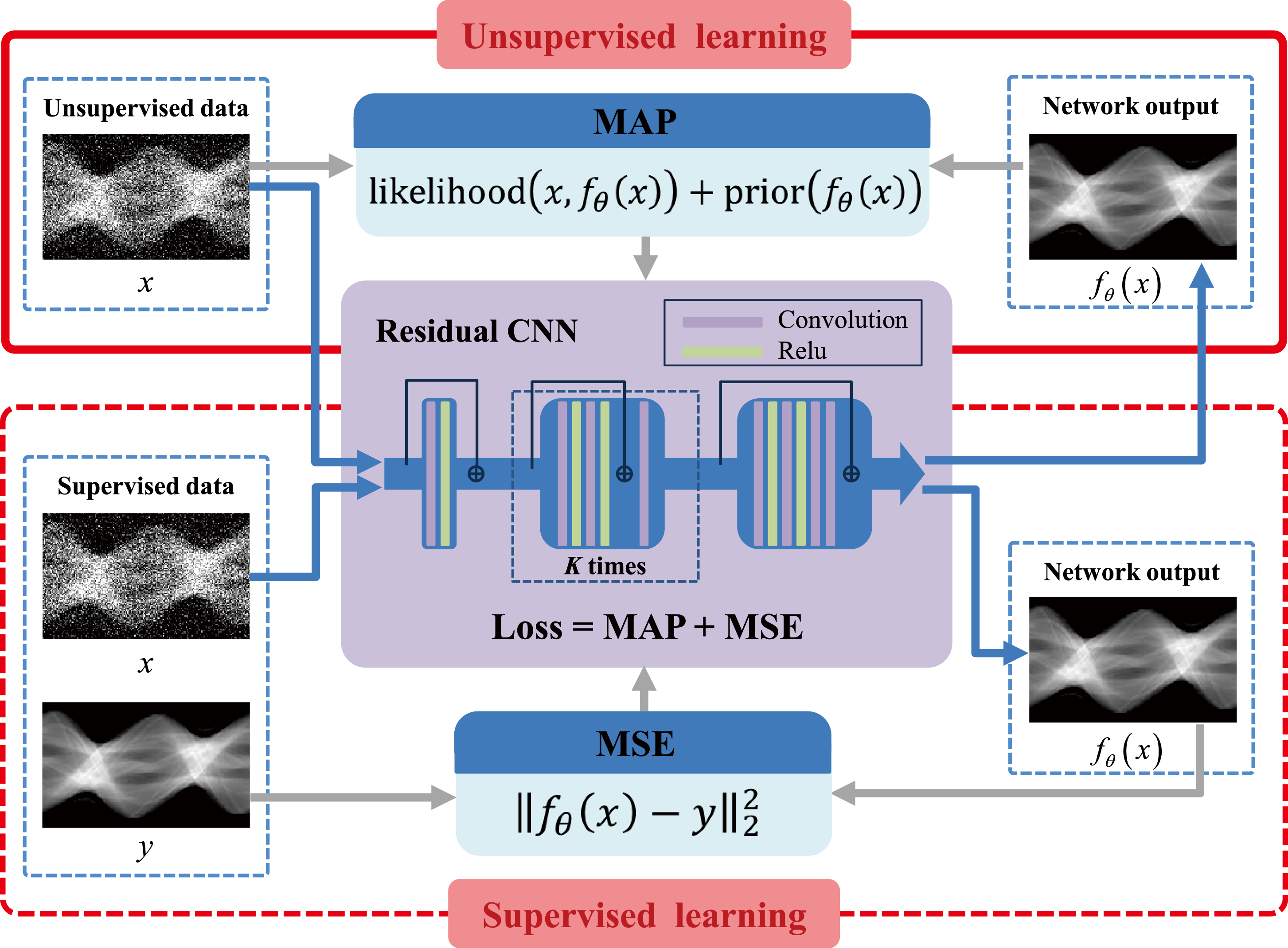}
		\vspace{-2mm}
		\caption{The framework of the proposed method. The unsupervised part (solid line) is trained directly on low-dose CTs, where the MAP model will guide the right gradients for network training. The supervised part (dashed line) is guided by the supervised low-dose/high-dose CT pairs, in traditional DL training manner. These two parts can be combined to form a semi-supervised DL framework.}
		\label{Fig:1}
	\end{center}
	\vspace{-3mm}
\end{figure}

Against the aforementioned issue, this work presents an unsupervised DL regime for directly involving unlabeled LdCT sinograms, without requirement of their high-dose ones, into network training. Specifically, through fully exploring both the structure characteristics underlying a clean CT sinogram and specific noise configuration in a measured LdCT sinogram, we can use a MAP objective to fully represent both of these knowledge contained in LdCT sinograms by elaborately designed regularization and likelihood terms, respectively. Such a MAP model facilitates an effective exploration on the gradient direction along which the input LdCT sinogram should be oriented to the expected clean one, and thus can be readily employed into the network training. Such an unsupervised DL regime can also be easily integrated with supervised CT sinogram pairs to further ameliorate the performance of the method. The basic implementation mechanism of the method is illustrated in Fig. \ref{Fig:1}.

In summary, this paper mainly makes the following contributions:

(1) Towards the low-dose CT enhancement issue, this work first proposes a feasible unsupervised DL regime, without need of supervised low-dose/high-dose CT sinogram pairs as inputs/outputs of the network, while directly being implemented on unsupervised LdCT sinograms. This method facilitates a sufficient utilization of unlabeled LdCT sinograms, and makes the DL strategy capable of being more easily and generally implemented in real scenarios with few high-dose data sources.

(2) We further extend our unsupervised DL method to semi-supervised version. For supervised data, we construct the objective function in the data-driven manner according to its supervised information. For unsupervised data, we construct the objective function in the model-driven manner by fully considering its prior structure and noise configuration. Such supervised and unsupervised integration is also inspiring to construct more general semi-supervised DL paradigms for other tasks.

(3) We have verified the superiority of the proposed unsupervised/semi-supervised DL strategy on real LdCT sinograms, in terms of both computational speed and accuracy, as compared with the traditional methods.

This paper is organized as follows: Section \ref{related_work} introduces some related works on the investigated task. Section \ref{model} presents our basic models and algorithms. Section \ref{experiments} demonstrates the experimental results and finally we give the conclusion in Section \ref{conclusion}.

\vspace{-3mm}
\section{Related Work}\label{related_work}\vspace{-2mm}
\subsection{Traditional Low-dose CT enhancement approaches}\vspace{-1mm}
Traditional methods can be mainly categorized into two classes, sinogram statistical iterative methods, which only use the information in sinogram domain, and  model-based iterative reconstruction (MBIR) methods, which combine the information of sinogram domain and CT image domain.

The penalized weighted least-square (PWLS) method is the representative work on the first class. One typical PWLS method was proposed by Wang et al. \cite{Wang2009Iterative}, who modeled accurate noise distribution and imposed a proper regularization to reduce sinogram noise. Moreover, Xie et al. \cite{Xie2017Robust} proposed a method taking full use of both the statistical properties of projection data and prior structure knowledge under sinogram domain for CT denoising and reached the state-of-the-art in this kind of methods. Comparatively, MBIR methods can offer the potential to reconstruct CT image with better bias-variance performance by using prior information of CT image domain. Some works explored different prior information in recovery model, such as total variation(TV) and its variants \cite{Bouman1993A}\cite{Tian2011Low}\cite{Zhu2010Duality}, dictionary learning \cite{Xu2014Dictionary} and nonlocal means \cite{Chen2009Bayesian}.

Though some of these methods show satisfactory effects on certain LdCT images, they can only be implemented on each CT image separately, while cannot get a deterministic prediction function to directly input LdCT images and output expected clean ones. This makes them always very time-consuming in real scenarios. Besides, such methods can only make use of one CT image to explore its latent ground-truth one, while cannot integrate more CT images to summarize its insightful statistical common knowledge and serve such useful knowledge for further LdCT enhancement. DL techniques thus attract more attention recently by finely ameliorating these issues.

\subsection{Deep learning approaches}\vspace{-1mm}
Currently, the data-based DL approaches has achieved inspiring achievements to this issue. For example, Chen et al. \cite{Chen2016Low} first introduced CNN in CT images denoising task. To extract features more efficiently, he further used a encoder-decoder network instead \cite{Chen2017Low}. After that, Yang et al. \cite{Yang2017CT} introduced the perceptual loss in CT enhancement task, which measured the difference between low-dose/high-dose CT image pairs in a high-level feature space to make them look more similar. Further, in order to preserve the detail information as well, Yi et al. \cite{Yi2018Sharpness} combined the GAN and a low-level feature space measurement network named sharpness detection network, to decrease the blur effect.

Though DL methods have an exciting performance and fast processing speed, they seriously rely on the pre-collected LdCTs and corresponding high-dose ones as their ground-truth. Such supervised samples, however, are always very hard to get and need take large costs including human labor and collecting time. Besides, the collection of high-dose CT images will always cause great harm to the patients' health. It is thus highly expected to have an unsupervised DL paradigm, by only inputting LdCTs into network for training without need of their guided high-dose ones.

\vspace{-0mm}
\section{Unsupervised/Semi-supervised Deep Learning for LdCT Enhancement}\label{model}\vspace{-2mm}
\subsection{Data-based supervised deep learning}\vspace{-1mm}

Let $x_{i}\in\mathbb{R}^{m}$, $i=1,2,...,n$ be the LdCT sinogram and $y_{i}\in\mathbb{R}^{m}$ be the corresponding high-dose sinogram, $f_{\theta}$ is the mapping $f_{\theta}$:$x\rightarrow{\widetilde{y}}$, where $\widetilde{y}$ is the predicted output by CNN. We thus denote $\widetilde{y}$ as $f_{\theta}(x)$, where $\theta$ is the parameters of CNN mapping. In the DL network for CT image enhancement, training dataset is a set of input-target pairs $\{x_{i}, y_{i}\}_{i=1,...,n}$. A commonly utilized strategy is to minimize the following mean square error (MSE) for network parameter tuning:
\begin{equation}
\min_{\theta}\sum\nolimits_{i}L(f_{\theta}(x_{i}),y_{i})=\min_{\theta}\sum\nolimits_{i}\|f_{\theta}(x_{i})-y{_{i}}\|_{2}^{2}.
\end{equation}

The supervised DL model can achieve a good trade-off between image quality and computational efficiency under adequately corrected supervised CT sinogram pairs. However, the number of the supervised sinograms would affect
the network's capability and sometimes the ground-truth sinograms are difficult to obtain. Instead, the unlabeled LdCT sinograms can be relatively easily collected. Therefore, the unlabeled LdCT dataset is expected to be also used in the network training to further improve enhancement performance.

\vspace{-2mm}
\subsection{Model-based unsupervised deep learning}\vspace{-1mm}
\label{gen_inst}

\begin{figure}\vspace{-4mm}
	\begin{center}
		\includegraphics[width=0.9\linewidth]{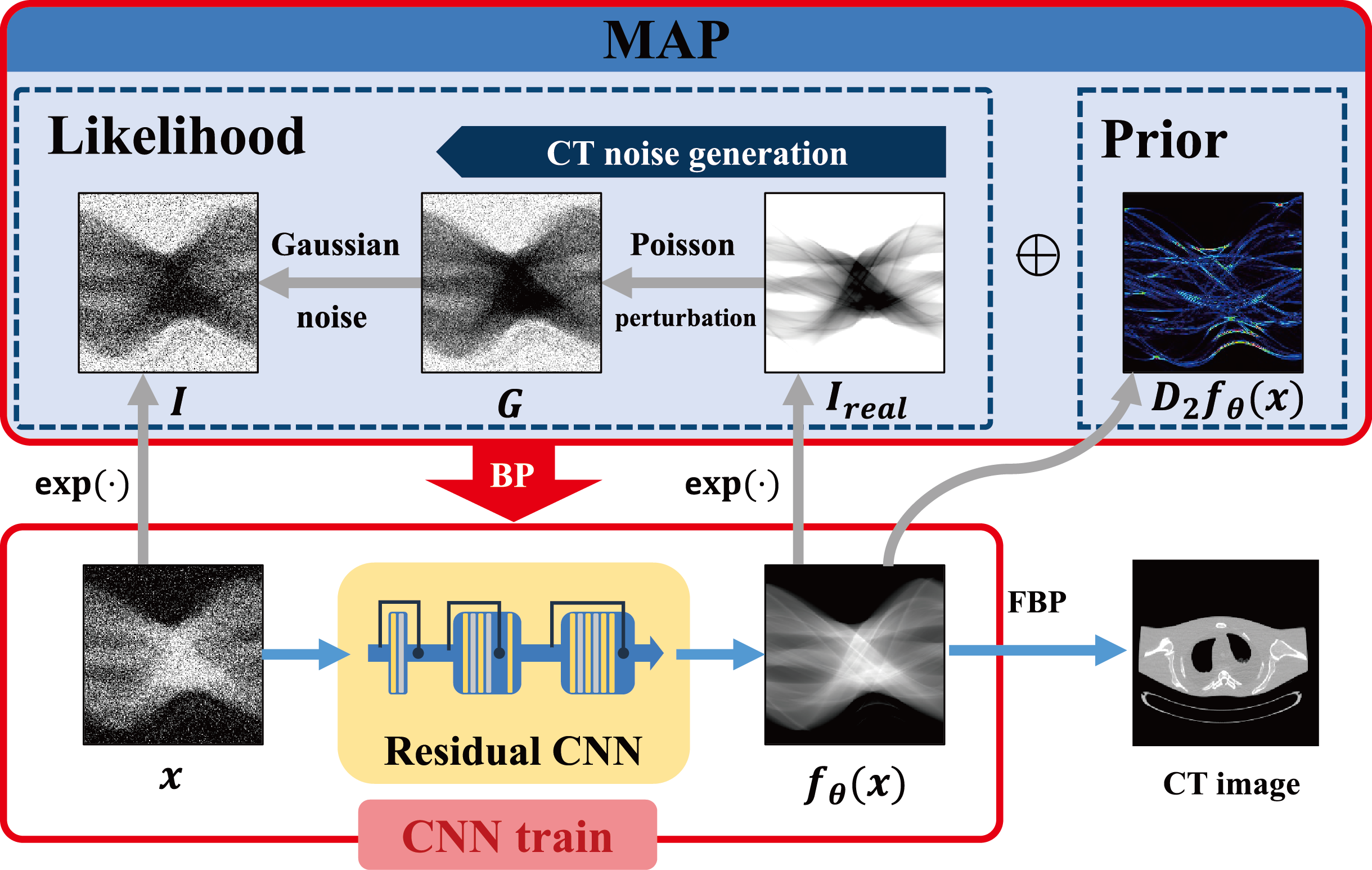}
		\vspace{-2mm}
		\caption{Illustration of the MAP model construction.}
		\label{Fig:2}
		\vspace{-3mm}
	\end{center}
\end{figure}

We first shortly introduce the basic generation process of a CT image.
$I_{0}$ denotes the number of unattenuated photons (X-ray fluence), $I$ is the number of the attenuated photons arriving at the detector and usually follows the combined Poisson-Gaussian noise distribution. Specifically, based on \cite{Xie2017Robust}, $G$ represents the atteunated photons with additive electronic noise only, ignoring the quanta fluctuation of X-ray interactions theoretically. And $I_{real}$ denotes the attenuated photons without any noise, which can be considered as the desired projection data. CT sinogram can be generated by $I$ after a logarithmic transformation. The final CT image can be got by FBP. The overall generation process is illustrated in Fig. \ref{Fig:2}.

We then introduce how to express the statistical properties (leading to noises), as well as its expected recovery structures, underlying a LdCT sinogram. This knowledge is then used to achieve a proper gradient direction to feed into the network and make the input LdCT sinogram orient to the expected clean ones as well as remove the unexpected noise.

The projection data $I$ are generally mixed with noise and can be expressed as follows:
\begin{equation}
I=G+\varepsilon,
\end{equation}
where $\varepsilon\in\mathbb{R}^{N}$ denotes the electronic noise. For projection $I$, the first term follows X-ray photon statistics and the second term leads to electronic noise background \cite{Riviere2004Reduction}\cite{Xu2009Electronic}. The electronic noise $\varepsilon$ can be described as a simple Gaussian distribution \cite{Ma2012Iterative}:
$\varepsilon_{j}\sim\mathcal N (\varepsilon_{j}|0,\sigma^2)$,
where $\sigma^2$ denotes the variance of noise. Based on \cite{Ma2012Iterative} we can obtain that
\begin{equation}\label{PIG}
p(I|G)=\frac{1}{(2\pi)^{N/2}\sigma^{N}}\exp^{-\frac{\|I-G\|_{2}^{2}}{2\sigma^{2}}}.
\end{equation}
The received quanta $G$ can be well depicted by the compound Poisson distribution \cite{La2006Penalized} \cite{Xie2017Robust} as follows:
\begin{equation}\label{PGF}
p(G|f_{\theta}(x))=\prod_{j=1}^{N}\left(\frac{({I_{0}}_{j}e^{-f_{\theta}(x)_{j}})^{G_{j}}}{G_{j}!}\exp(-I_{0j}e^{-f_{\theta}(x)_{j}}))\right),
\end{equation}
where ${I_{0}}_{j}$ denotes $I_0$ along the projection path $j$. $x$ and $f_{\theta}(x)$ denote the log-transformations of $I$ and $I_{real}$, corresponding to the input LdCT sinogram and ideal output one of the network, respectively, with elements $(x)_{j}$ and $f_{\theta}(x)_{j}$, where $N$ is the total number of measurements in the scan. By combining (\ref{PIG}) and (\ref{PGF}), the generative-model of projection data can be obtained as \cite{Xie2017Robust}:
\begin{equation}\label{PIG_F}
p(I,G|f_{\theta}(x))=\prod_{j=1}^{N}\left(\frac{({I_{0}}_{j}e^{-f_{\theta}(x)_{j}})^{G_{j}}}{G_{j}!}\exp(-{I_{0}}_{j}e^{-f_{\theta}(x)_{j}}))\right)\frac{1}{(2\pi)^{N/2}\sigma^{N}}\exp^{-\frac{\|I-G\|_{2}^{2}}{2\sigma^{2}}}.
\end{equation}
Note that the electronic noise background and the statistical property of photon statistics have been considered in (\ref{PIG}) and (\ref{PGF}). Besides the above understanding on the statistical properties on LdCT sinogram, constituting the likelihood term in the optimized MAP model, we can also get useful prior knowledge on the ideal recovery, to further compensate the model.

Based on \cite{Xie2017Robust}, the sinogram data is formed as a manifold approximately
constituted by a combination of several flat surface, This flats-combination prior can be introduced to describe the properties of the sinogram, i.e., sparsity in its second order derivative, which can be formulated as:
\begin{equation}\label{PF}
p(f_{\theta}(x))\propto e^{-k\|\ln(D_{2}f_{\theta}(x)+\varepsilon)-\ln(\varepsilon)\|_1},
\end{equation}
where $k$ is a constant parameter, and $D_{2}$ is the second order difference matrix. This class of distribution can encode the transformed-sparsity of data.

By combining (\ref{PIG_F}) and (\ref{PF}), the complete posterior distribution on data can be formulated as follows:
\small
\begin{equation}
	\begin{split}
	p(G,f_{\theta}(x)|I)&=\frac{p(I,G|f_{\theta}(x))p(f_{\theta}(x))}{p(I)}\\
	&\propto\exp\left(-\frac{\|I-G\|_{2}^{2}}{2\sigma^{2}}-k\|\ln(D_{2}f_{\theta}(x)+\varepsilon)-\ln(\varepsilon)\|_1\right)\prod_{j=1}^{N}\left(\frac{({I_{0}}_{j}e^{-f_{\theta}(x)_{j}})^{G_{j}}}{G_{j}!}\right).
	\end{split}
\end{equation}
\normalsize
%$$p(G,f_{\theta}(x_{i})|I)=\frac{p(I,G|f_{\theta}(x_{i}))p(f_{\theta}(x_{i}))}{p(I)}$$
%\begin{equation}
%\propto\exp\left(-\frac{\|I-G\|_{2}^{2}}{2\sigma^{2}}-k\|\ln(D_{2}f_{\theta}(x_{i})+\varepsilon)-\ln(\varepsilon)\|_1\right)\prod_{j=1}^{N}\left(\frac{({I_{0}}_{j}e^{-f_{\theta}(x_{i})_{j}})^{G_{j}}}{G_{j}!}\right).
%\end{equation}
The ideal $f_{\theta}(x)$ can be estimated under MAP framework \cite{Fessler2000Statistical}\cite{La2006Penalized}\cite{Ma2012Iterative}. The network can then be trained under the guidance of the following loss term in an unsupervised DL manner:
\begin{equation}
	\begin{split}
	&\mathop{\arg\min}_{\theta,G}\sum\nolimits_{i}\sum\nolimits_{j=1}^{N}\left(\frac{\|I-G\|_{2}^{2}}{2\sigma^{2}}-G_{i}\ln({I_{0}}_{j})+G_{j}f_{\theta}(x_{i})_{j}+\ln(G_{j}!)+{I_{0}}_{j}e^{f_{\theta}(x_{i})_{j}}\right)\\
	&~~~~~~~+k\sum\nolimits_{i}\|\ln(D_{2}f_{\theta}(x_{i})+\varepsilon)-\ln(\varepsilon)\|_{1}.
	\end{split}
\end{equation}
%$$\min_{\theta,G}\sum\nolimits_{i}\sum\nolimits_{j=1}^{N}\left(\frac{\|I-G\|_{2}^{2}}{2\sigma^{2}}-G_{i}\ln({I_{0}}_{j})+G_{j}f_{\theta}(x_{i})_{j}+\ln(G_{j}!)+{I_{0}}_{j}e^{f_{\theta}(x_{i})_{j}}\right)$$
%\begin{equation}
%+k\|\ln(D_{2}f_{\theta}(x_{j})+\varepsilon)-\ln(\varepsilon)\|.
%\end{equation}

\vspace{-2mm}
\subsection{Semi-supervised deep learning}\vspace{-1mm}
We can then naturally construct semi-supervised DL scheme to fully utilize both supervised and unsupervised LdCT data sources, via combining the aforementioned two types of models. The corresponding loss function can be expressed as follows:
\small
\begin{equation}\label{objective}
	\begin{split}
	&L(G,\theta)=\sum\nolimits_{x_{i}\in{C_{1}}}\bigg(\sum\nolimits_{j}\Big(\frac{\|I-G\|_{2}^{2}}{2\sigma^{2}}-G_{j}\big(\ln{{I_{0}}_{j}}\big)+G_{j}f_{\theta}(x_{i})_{j}+\ln{G_{j}!}+{I_{0}}_{j}e^{-f_{\theta}(x_{i})_{j}}\Big)\\
	&~~~~~~~~~~~~~~~~~~+k\|\ln(D_{2}f_{\theta}(x_{i})+\varepsilon)-\ln(\varepsilon)\|_{1}\bigg)+\lambda\sum\nolimits_{x_{i}\in{C_{2}}}\|f_{\theta}(x_{i})-y_{i}\|_{2}^{2},
	\end{split}
\end{equation}
\normalsize
%$$L(G,\theta)=\sum\nolimits_{x_{i}\in{C_{1}}}\bigg(\sum\nolimits_{j}\Big(\frac{\|I-G\|_{2}^{2}}{2\sigma^{2}}-G_{j}\big(\ln{{I_{0}}_{j}}\big)+G_{j}f_{\theta}(x_{i})_{j}+\ln{G_{j}!}+{I_{0}}_{j}e^{-f_{\theta}(x_{i})_{j}}\Big)$$
%\begin{equation}
%+k\|\ln(D_{2}f_{\theta}(x_{i})+\varepsilon)-\ln(\varepsilon)\|_{1}\bigg)+\lambda\sum\nolimits_{x_{i}\in{C_{2}}}\|f_{\theta}(x_{i})-y_{i}\|_{2}^{2}\\.
%\end{equation}
where $C_{1}$ and $C_2$ are the sets of unsupervised and supervised sinogram data, respectively. $\lambda$ is the trade-off parameter, which balances the loss functions of supervised and unsupervised learning components. The value of this parameter can be set based on the portions of two parts of data. The more we have the supervised ones, the larger it should be set.

By using such loss setting, the network can be trained both on supervised and purely unsupervised inputs. Note that when $\lambda=0$, this model will directly degenerate to the unsupervised one.

\begin{algorithm}[tb]
  \caption{ Algorithm for Update $G$}\label{alg:update_g}
  \begin{algorithmic}[1]
    \Require
      $x_{i}$, $I$, $I_{0}$ and $\sigma$
    \Ensure
      $G$
    \State Initialize $G$ as the value of $G$ we obtained in the last step of complete algorithm
    \State \textbf{for} $j$ = 1 : $N$ \textbf{do}
    \State \quad \textbf{while} $h(G_{j})\textgreater h(G_{j}+1)$ \textbf{do}
    \State \qquad Update $G_{j}=G_{j}+1$
    \State \quad \textbf{end while}
    \State \quad \textbf{while} $h(G_{j})\textgreater h(G_{j}-1)$ \textbf{do}
    \State \qquad Update $G_{j}=G_{j}-1$
    \State \quad \textbf{end while}
    \State \textbf{end for}
  \end{algorithmic}
\end{algorithm}

\vspace{-2mm}
\subsection{Alternative optimization algorithm for solving the model}\label{algorithm}\vspace{-1mm}

We can readily employ the alternative optimization algorithm to calculate (\ref{objective}). The optimization procedures can be summarized as follows:

With the other parameters fixed, $G$ can be updated by solving $\mathop{\arg\min}_{G}L(G,\theta)$, that is:
\begin{equation}
\mathop{\arg\min}_{G}\sum\nolimits_{j}\left(\frac{\|I-G\|_{2}^{2}}{2\sigma^{2}}-G_{j}\left(\ln{{I_{0}}_{j}}\right)+G_{j}f_{\theta}(x_{i})_{j}+\ln{G_{j}!}\right).
\end{equation}
This problem can be further separated for each $G_{j}$ as:
\begin{equation}
\mathop{\arg\min}_{G}h\left(G_{j}\right)=\frac{\|I-G\|_{2}^{2}}{2\sigma^{2}}-G_{j}\left(\ln{{I_{0}}_{j}}\right)+G_{j}f_{\theta}(x_{i})_{j}+\ln{G_{j}!},
\end{equation}
whose solution can be obtained by Algorithm \ref{alg:update_g}.

With the other parameters fixed, $\theta$ can be updated by solving $\mathop{\arg\min}_{\theta}L(G,\theta)$, which is equivalent to the following problem:
\begin{equation}
	\begin{split}
	&\mathop{\arg\min}_{\theta}\sum\nolimits_{x_{i}\in{C_{1}}}\left(\sum\nolimits_{j}\left(G_{j}f_{\theta}(x_{i})_{j}+{I_{0}}_{j}e^{-f_{\theta}(x_{i})_{j}}\right)+k\|\ln(D_{2}f_{\theta}(x_{i})+\varepsilon)-\ln(\varepsilon)\|_{1}\right)\\
	&~~~~~~~~~+\lambda\sum\nolimits_{x_{i}\in{C_{2}}}\|f_{\theta}(x_{i})-y_{i}\|_{2}^{2}.
	\end{split}
\end{equation}
%$$\min_{\theta}\sum_{x_{i}\in{C_{1}}}\left(\sum_{j}\left(G_{j}f_{\theta}(x_{i})_{j}+{I_{0}}_{j}e^{-f_{\theta}(x_{i})_{j}}\right)+k\|\ln(D_{2}f_{\theta}(x_{i})+\varepsilon)-\ln(\varepsilon)\|\right)$$
%\begin{equation}
%+\lambda\sum_{x_{i}\in{C_{2}}}\|f_{\theta}(x_{i})-y_{i}\|_{2}^{2}.
%\end{equation}
This corresponds to a standard network training problem, and can be easily calculated by calling any of current deep learning algorithms, like Adam \cite{Kingma2014Adam}, for network parameter tuning.

The whole procedure for optimization of (\ref{objective}) can be summarized in Algorithm \ref{alg:all}.

\begin{algorithm}[tb]
  \caption{ Algorithm for Solving (\ref{objective})}\label{alg:all}
  \begin{algorithmic}[1]
    \Require
      $I_{0}$, $I=I_{0}\odot{e^{-x_{i}}}$
    \Ensure
      the denoised scan $f_{\theta}(x)$
    \State Initialize CNN parameter $\theta$
    \State \textbf{while} not convergence \textbf{do}
    \State \quad Update $G$ by Algorithm \ref{alg:update_g}.
    \State \quad Update $\theta$ by Back Propagation Algorithm.
    \State \textbf{end while}
  \end{algorithmic}
\end{algorithm}

\vspace{-0mm}
\section{Experimental results}\label{experiments}\vspace{-2mm}
The performance of the proposed two methods, i.e., unsupervied learning (unsup-CNN) and semi-supervised learning (semi-CNN), is verified in this section. Comparison methods include PWLS \cite{Wang2009Iterative}, MAP-FC \cite{Xie2017Robust}, and supervised CNN methods. The PSNR, SSIM \cite{Wang2004Image} and FSIM \cite{Zhang2011FSIM} are used for performance evaluation. In addition, the running time of all algorithms are demonstrated for speed comparison. The \emph{"2016 Low-dose CT Grand Challenge datasets"}\footnote{\url{https://www.aapm.org/GrandChallenge/LowDoseCT}} were used in the experiments. The normal-dose CT data were acquired with 120 kVp and 200 effective mAs. LdCT data at three dose levels, 20 mAs, 12.5 mAs and 10 mAs, were generated via the simulation method \cite{Zeng2015A}. The high-dose ones at 200 mAs are considered as the ground-truth for comparison in the experiments. More detailed information on the utilized network settings can be referred to in supplementary material.

\vspace{-2mm}
\subsection{On the effect of unsupervised DL method}\vspace{-1mm}
\begin{figure}\vspace{-1mm}
	\begin{center}
		\includegraphics[width=0.9\linewidth]{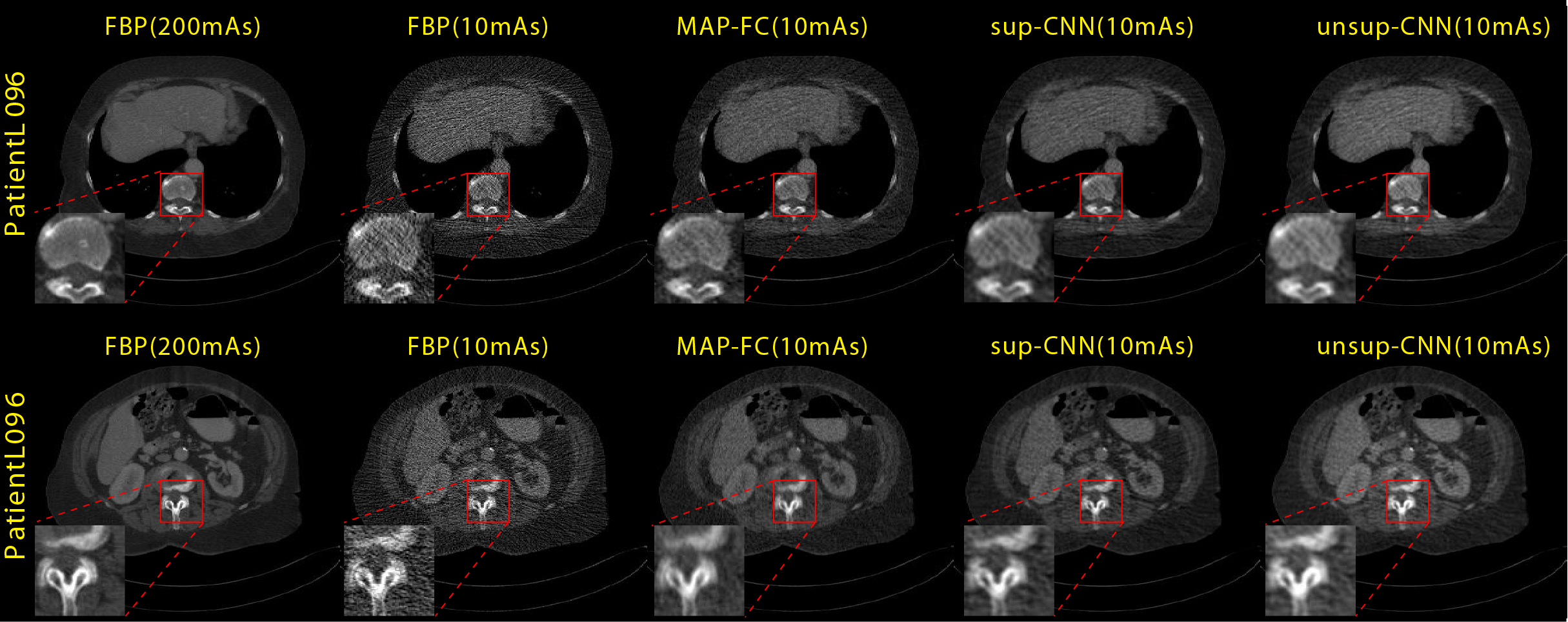}\vspace{-2mm}
		\caption{The high-dose sinogram (200 mAs) and low-dose images processed by the FBP, MAP-FC, sup-CNN and unsup-CNN methods. All the images are displayed in the same window.}
		\label{Fig:exp1}
	\end{center}\vspace{-2mm}
\end{figure}
To verify the effectiveness of the unsupervised CNN (unsup-CNN) method, supervised CNN (sup-CNN) method and MAP-FC method were conducted for comparison. For unsupervised CNN, we only used 50 LdCT sinograms as training set. For sup-CNN method, we used 50 low-dose/high-dose CT sinogram pairs as training set.

\begin{table}[tbp]\vspace{-0mm}
  \centering\small
   \caption{Quantitative measurements of results of all competing methods at three different noise levels.}\label{tab:1}\vspace{-1mm}
    \begin{tabular}{rlrrrrrrr}
    \toprule
    \multicolumn{1}{l}{Patient} &       &       & \multicolumn{1}{l}{L096} &       &       & \multicolumn{1}{l}{L096} &       &  \\
\cmidrule{3-9}    \multicolumn{1}{l}{dose} &       & \multicolumn{1}{l}{PSNR} & \multicolumn{1}{l}{SSIM} & \multicolumn{1}{l}{FSIM} & \multicolumn{1}{l}{PSNR} & \multicolumn{1}{l}{SSIM} & \multicolumn{1}{l}{FSIM} & \multicolumn{1}{l}{Time} \\
    \midrule
    \multicolumn{1}{l}{10mAs} & FBP & 29.2551 & 0.6157 & 0.8916 & 33.2117 & 0.7662 & 0.9274 & - \\
          & MAP-FC & 35.3356 & 0.8465 & 0.9461 & 38.9901 & 0.9224 & 0.9681 & 21.76 \\
          & sup-CNN & \textbf{36.8963} & \textbf{0.9128} & \textbf{0.9658} & 39.1643 &\textbf{ 0.9494} & 0.9738 &\textbf{ 0.083} \\
          & unsup-CNN & 35.5475 &\textbf{ 0.9128} & 0.9629 & \textbf{40.1458} & 0.9471 &\textbf{ 0.9745} & \textbf{0.083} \\
    \midrule
    \multicolumn{1}{l}{12.5mAs} & FBP & 30.4716 & 0.6799 & 0.9178 & 34.5637 & 0.8111 & 0.9444 & - \\
          & MAP-FC & 36.5699 & 0.8786 & 0.9602 & 39.9781 & 0.9376 & 0.9762 & 21.76 \\
          & sup-CNN & \textbf{37.6384} & 0.9244 & 0.9725 & 40.0542 & \textbf{0.9622} & \textbf{0.9823} & \textbf{0.083} \\
          & unsup-CNN & 36.1274 & \textbf{0.9338} &\textbf{ 0.9974} & \textbf{40.3582} & 0.9475 & 0.9803 & \textbf{0.083} \\
    \midrule
    \multicolumn{1}{l}{20mAs} & FBP & 33.6759 & 0.7887 & 0.9521 & 37.3508 & 0.8851 & 0.9682 & - \\
          & MAP-FC & 38.7972 & 0.9229 & 0.9771 & 41.7376 & 0.9593 & 0.9841 & 21.76 \\
          & sup-CNN & \textbf{38.8859} & 0.9406 & 0.9812 & \textbf{42.1377} & \textbf{0.9593} & \textbf{0.9855} & \textbf{0.083} \\
          & unsup-CNN & 38.3038 & \textbf{0.9439} & \textbf{0.9813} & 41.3767 & 0.9593 & 0.9846 & \textbf{0.083} \\
    \bottomrule
    \end{tabular}\vspace{-2mm}
\end{table}%

Fig. \ref{Fig:exp1} shows the corresponding results processed by the different methods. It can be observed that MAP-FC, sup-CNN and unsup-CNN can suppress noises effectively. Because the bone regions contain abundant structures details, two regions indicated by the red boxes are selected to validate image quality improvement. It is seen that the unsup-CNN method preserves more details with higher resolution in the magnified ROI than the other competing methods. In addition, Table \ref{tab:1} lists the PSNR, FSIM and SSIM measurements and running time of all competing methods. It is seen that the two CNN-based methods perform better than the MAP-FC method in all cases. And the unsup-CNN method can obtain similar performance to the sup-CNN method, while the latter require extra supervised samples for training. These results substantiate that the proposed unsup-CNN method can properly extract gradients to guide network training on purely unsupervised LdCT sinograms.

\vspace{-2mm}
\subsection{On the effect of unsupervised DL method for single image}\vspace{-1mm}

\begin{figure}[t]\vspace{-0mm}
	\begin{center}
		\includegraphics[width=0.9\linewidth]{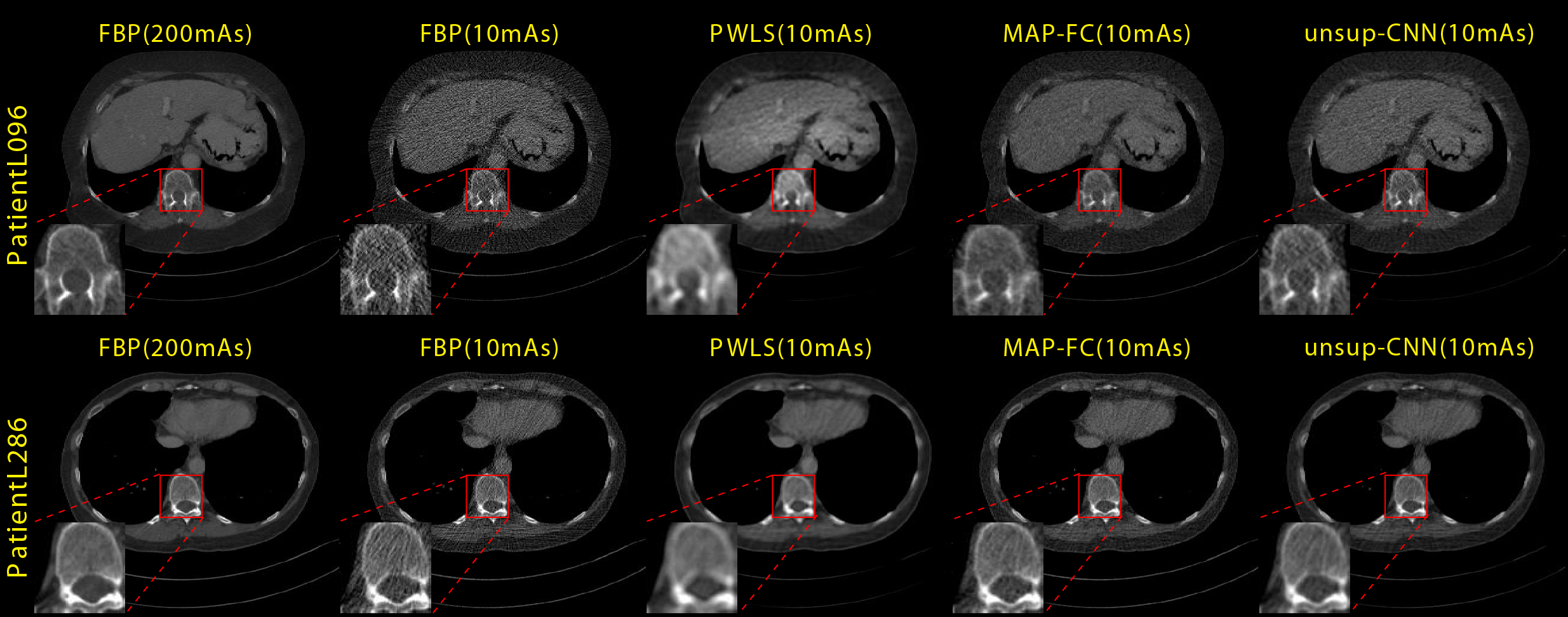}\vspace{-2mm}
		\caption{The high-dose sinogram (200 mAs) and low-dose sinogram measurements processed by the FBP, PWLS, MAP-FC, sup-CNN and unsup-CNN methods. All the images are displayed in the same window.}
		\label{Fig:exp2}
	\end{center}\vspace{-2mm}
\end{figure}

To evaluate the capability of the unsupervised network in extreme cases, only one LdCT sinogram was used for network training. Fig. \ref{Fig:exp2} shows the CT images processed by the FBP, PWLS, MAP-FC and unsup-CNN methods. Through visual inspection, it is seen that the PWLS and MAP-FC can suppress noise-induced artifacts well at the cost of resolution loss. The proposed unsup-CNN method effectively reduces the noise-induced artifacts and also preserves the resolution successfully.

This experiment illustrates that the proposed unsup-CNN can also work well even the training data is limited. This is attributed to that network can also work as an optimizer to minimize the target function. Note that in such scenarios, one superiority of the proposed method is that it can get a explicit prediction network, which can be efficiently utilized for further CT enhancement task.

\begin{figure}\vspace{-0mm}
	\begin{center}
		\includegraphics[width=0.9\linewidth]{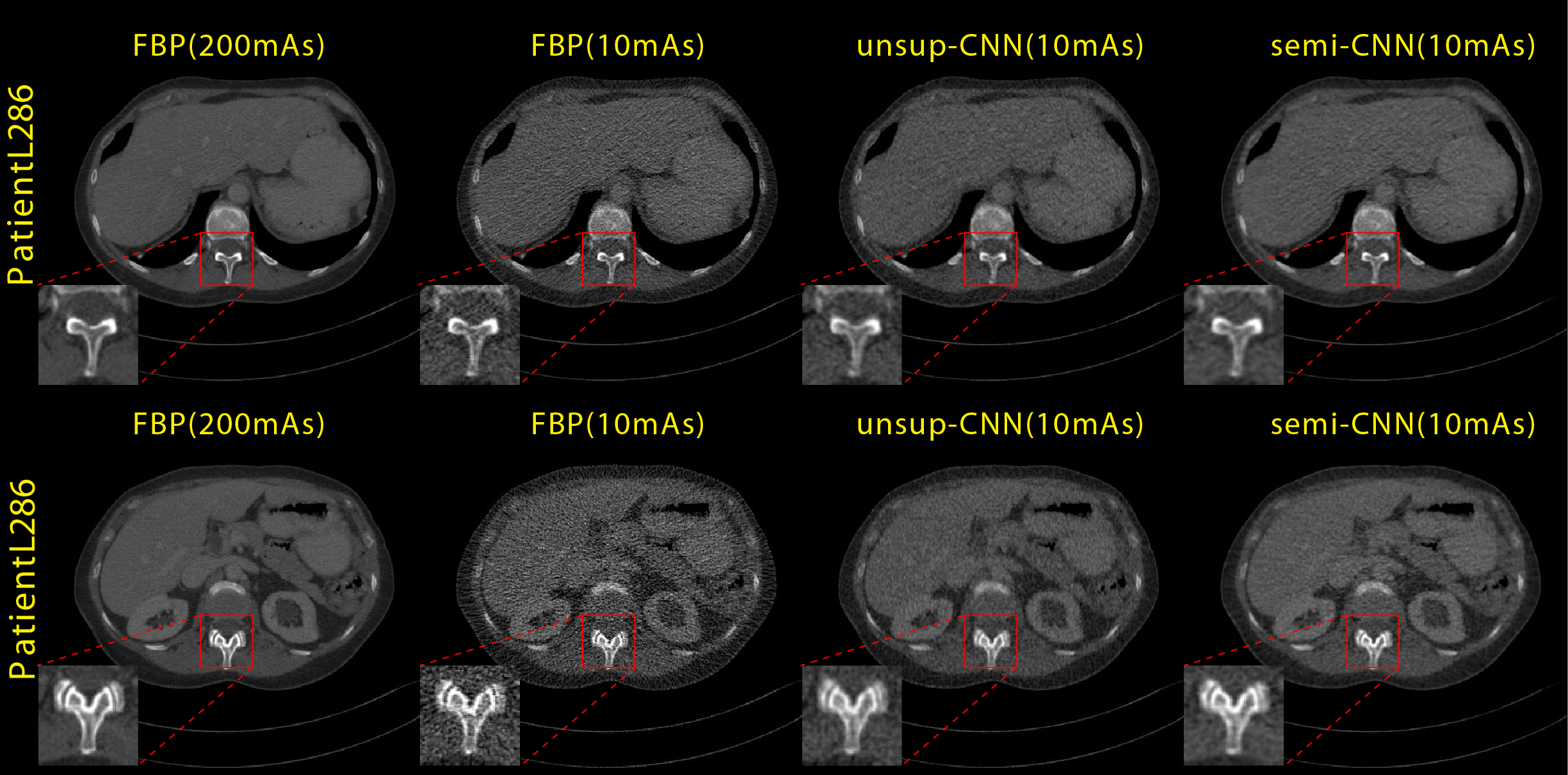}\vspace{-2mm}
		\caption{The high-dose sinogram (200 mAs) and low-dose sinogram measurements processed by the FBP, sup-CNN and semi-CNN methods. All the images are displayed in the same window.}
		\label{Fig:exp3}
	\end{center}\vspace{-2mm}
\end{figure}

\vspace{-2mm}
\subsection{On the effect of semi-supervised DL method}\vspace{-1mm}
We used 20 low-dose/high-dose CT sinogram pairs (supervised samples) and 50 LdCT sinograms (unsupervised samples) as training data in the semi-supervised CNN experiment. For comparison, we used the same LdCT data as training data in the unsupervised CNN network. Fig. \ref{Fig:exp3} shows the LdCT results processed by the FBP, unsup-CNN and semi-CNN methods. It is evident that both of the proposed CNN-based methods are able to remove noise-induced artifacts satisfactorily compared to the high-dose (200 mAs) one. The semi-CNN method performs better than the unsup-CNN method in the noise-induced artifacts in the flat region. The zoomed bone regions (ROIs indicated by the red boxes) suggests that the proposed semi-CNN method can reconstruct the fine structures with higher resolution than the unsup-CNN method.

Table \ref{tab:2} lists the PSNR, FSIM, and SSIM measurements for the results with the FBP, unsup-CNN and semi-CNN methods at three noise levels. Both the CNN-based methods outperform the conventional FBP method. The semi-CNN method that combinationally uses the supervised and unsupervised CT data sources leads to its better reconstruction quality than the unsup-CNN methods.

\begin{table}[tbp]
  \centering
  \caption{The quantitative measurements of the results with the different methods at three noise levels.}\vspace{-1mm}\label{tab:2}\small
     \begin{tabular}{rlrrrrrr}
    \toprule
    \multicolumn{1}{l}{Patient} &       &       & \multicolumn{1}{l}{L286} &       &       & \multicolumn{1}{l}{L286} &  \\
\cmidrule{3-8}    \multicolumn{1}{l}{dose} &       & \multicolumn{1}{l}{PSNR} & \multicolumn{1}{l}{SSIM} & \multicolumn{1}{l}{FSIM} & \multicolumn{1}{l}{PSNR} & \multicolumn{1}{l}{SSIM} & \multicolumn{1}{l}{FSIM} \\
    \midrule
    \multicolumn{1}{l}{10mAs} & FBP & 29.9772 & 0.6426 & 0.8857 & 29.8674 & 0.6275 & 0.8757 \\
          & unsup-CNN & 36.6898 & 0.9053 & 0.9514 & 36.7192 & 0.9048 & 0.9515 \\
          & semi-CNN & \textbf{37.4342} & \textbf{0.9165} & \textbf{0.9604} & \textbf{37.6782} & \textbf{0.9569} & \textbf{0.9592} \\
    \midrule
    \multicolumn{1}{l}{12.5mAs} & FBP & 31.4157 & 0.7042 & 0.9137 & 31.2056 & 0.6875 & 0.9002 \\
          & unsup-CNN & 37.0445 & 0.9191 & 0.9649 & 36.9405 & 0.9224 & 0.9599 \\
          & semi-CNN & \textbf{38.2185} & \textbf{0.9128} & \textbf{0.9681} & \textbf{38.2185} & \textbf{0.9213} & \textbf{0.9637} \\
    \midrule
    \multicolumn{1}{l}{20mAs} & FBP & 34.3839 & 0.8088 & 0.9593 & 34.0601 & 0.7967 & 0.9401 \\
          & unsup-CNN & 38.4932 & 0.9384 & 0.9699 & 38.3164 & 0.9352 & 0.9378 \\
          & semi-CNN & \textbf{39.6660} & \textbf{0.9406} & \textbf{0.9776} & \textbf{39.5755} & \textbf{0.9457} & \textbf{0.9751} \\
    \bottomrule
    \end{tabular}\vspace{-3mm}
\end{table}%

\vspace{-3mm}
\section{Conclusion}\label{conclusion}\vspace{-2mm}

In this study, we have proposed a new mechanism on making DL performable on unsupervised training data, and especially realized it for the task of LdCT enhancement. Through sufficiently understanding and formulating the statistics properties embedded in data and prior structures underlying the expected recovery data, we can construct a MAP model, which facilitates an effective gradient direction to guide the unsupervised LdCT transformed to the expected clean one. Such gradients can easily feed into the network for its parameter tuning, and thus the deep learning can be implemented in an unsupervised manner. The proposed method constitutes a general paradigm to realize unsupervised/semi-supervised DL for more related tasks, like signal recovery and image reconstruction.

%\newpage

%\bibliographystyle{plain}
%\bibliography{ref1}

\end{document}